\PassOptionsToPackage{HTML,table,dvipsnames}{xcolor}

\documentclass[]{antgroup}
\usepackage{antgroup}

\usepackage{amsmath}
\usepackage{cleveref}
\usepackage{multicol}
\usepackage{amssymb}
\usepackage{array}
\usepackage{tabularx}
\usepackage{bbm}
\usepackage{makecell}
\usepackage{wrapfig}
\usepackage{colortbl}
\usepackage{threeparttable}
\usepackage[normalem]{ulem}
\usepackage{caption}
\tcbuselibrary{breakable,listings}

\useunder{\uline}{\ul}{}

\usepackage{amsmath,amsfonts,bm}

\def\eqref#1{equation~\ref{#1}}

\def\1{\bm{1}}

\DeclareMathAlphabet{\mathsfit}{\encodingdefault}{\sfdefault}{m}{sl}
\SetMathAlphabet{\mathsfit}{bold}{\encodingdefault}{\sfdefault}{bx}{n}

\newcommand*\justify{%
  \fontdimen2\font=0.4em
  \fontdimen3\font=0.2em
  \fontdimen4\font=0.1em
  \fontdimen7\font=0.1em
  \hyphenchar\font=`\-
}

\renewcommand{\texttt}[1]{%
  \begingroup
  \ttfamily
  \begingroup\lccode`~=`/\lowercase{\endgroup\def~}{/\discretionary{}{}{}}%
  \begingroup\lccode`~=`[\lowercase{\endgroup\def~}{[\discretionary{}{}{}}%
  \begingroup\lccode`~=`.\lowercase{\endgroup\def~}{.\discretionary{}{}{}}%
  \catcode`/=\active\catcode`[=\active\catcode`.=\active
  \justify\scantokens{#1\noexpand}%
  \endgroup
}

\usepackage{amsmath}
\usepackage{amssymb}
\usepackage{amsfonts}                               
\usepackage{amsthm}
\usepackage[mathcal]{eucal}
\usepackage{mathrsfs}
\usepackage{bm}                                     
\usepackage{blkarray}                               
\usepackage{nicefrac}                               

\usepackage{wrapfig}
\usepackage{graphicx}                               
\usepackage{caption}
\usepackage{cleveref}

\usepackage{tikz}                                          
\usepackage{circuitikz}
\usetikzlibrary{patterns,snakes}
\usetikzlibrary{positioning,calc,fit,decorations.pathmorphing,shapes.geometric, shapes.gates.logic.US, calc}
\usetikzlibrary{arrows,arrows.meta,decorations.markings,shapes,shapes.arrows}
\usetikzlibrary{decorations,decorations.pathreplacing}
\usetikzlibrary{backgrounds}
\usepackage{filecontents}                           
\usepackage{pgfplots}
\usepackage{pgfplotstable}
\usepgfplotslibrary{groupplots}
\usepackage{scalefnt}
\pgfplotsset{compat=newest}

\definecolor{LightGray}{gray}{0.9}
\definecolor{myPink}{HTML}{EE6B98}
\definecolor{myBlue}{HTML}{6BB8FA}
\definecolor{myPurple}{HTML}{9F63F0}
\definecolor{myBlueBase}{HTML}{147BFF}
\definecolor{myGreen}{HTML}{34A853}
\definecolor{myOrange}{HTML}{FF8C00}
\definecolor{hypCyan}{HTML}{00AFC7}

\hypersetup{
  pdftitle={OACap: Caption-Guided Active Observation for Detailed Video Captioning},
  pdfauthor={Hao Shi, Hong Yan, Xiaoyang Zhang, Chao Huang, Qing Cui, Jun Zhou},
  pdfsubject={Technical report},
  pdfkeywords={video captioning, caption-guided active observation, agent, benchmark, factuality}
}

\renewcommand\thefootnote{}

\title{OACap: Caption-Guided Active Observation\\
  for Detailed Video Captioning}

\author{\centering
Hao Shi$^{1,2*}$,
Hong Yan$^{1*}$,
Xiaoyang Zhang$^{1,3}$,
Chao Huang$^{1,\dag}$,
Qing Cui$^1$,
Jun Zhou$^{1,\ddag}$
}

\affiliation{$^1$Ant Group \quad $^2$Zhejiang University \quad $^3$Peking University}

\begin{document}
\maketitle

\footnotetext{$*$ Equal contribution; $\dag$ Team leader; $\ddag$ Corresponding author: jun.zhoujun@antgroup.com.}

\renewcommand\thefootnote{\arabic{footnote}}
\setcounter{footnote}{0}

\begin{abstract}
Unlike question answering, where a query specifies a perceptual target, open-ended detailed captioning must determine what evidence to acquire and how to integrate it into a coherent global description.
We propose \textbf{Observation-Adaptive Video Captioner (OACap)}, a caption-centered agent that turns unsupported or missing information into targeted observation requests, uses spatiotemporal tools to acquire evidence, and iteratively verifies and revises the caption.
OACap captions consistently improve performance across general video-understanding benchmarks using a fixed text-only reader.
To better diagnose the information they preserve, we introduce \textbf{Factual-Semantic Captioning Bench (FSCap-Bench)}.
Its \textit{Core--Support Factuality Score (CSFS)} combines claim precision with importance-aware fact coverage and responds as expected to four controlled perturbation types on $93.0$--$98.3\%$ of valid caption pairs.
Its \textit{Cross-Modal Semantic Round-Trip Score (CSRS)} measures semantic preservation through a video--caption--video--caption round trip by matching atomic facts from source and recovered captions.
Together, they connect downstream textual-memory utility with factual and semantic preservation.
\end{abstract}
\section{Introduction}

Detailed video captions must preserve entities, attributes, state changes, and temporal relations, yet a single global pass often misses small objects, brief actions, and distant transitions~\citep{sharegpt4video2024,videosalmonn2025}. Unlike VideoQA, open-ended captioning provides no query that specifies what evidence to seek. Text-only refinement cannot recover unseen facts, while existing tool-augmented video methods typically condition additional observation on an external query or reasoning state~\citep{wang2024videoagent,jeoung2024adaptive,avp2025}.

\begin{figure*}[!ht]
    \centering
    \includegraphics[width=\textwidth]{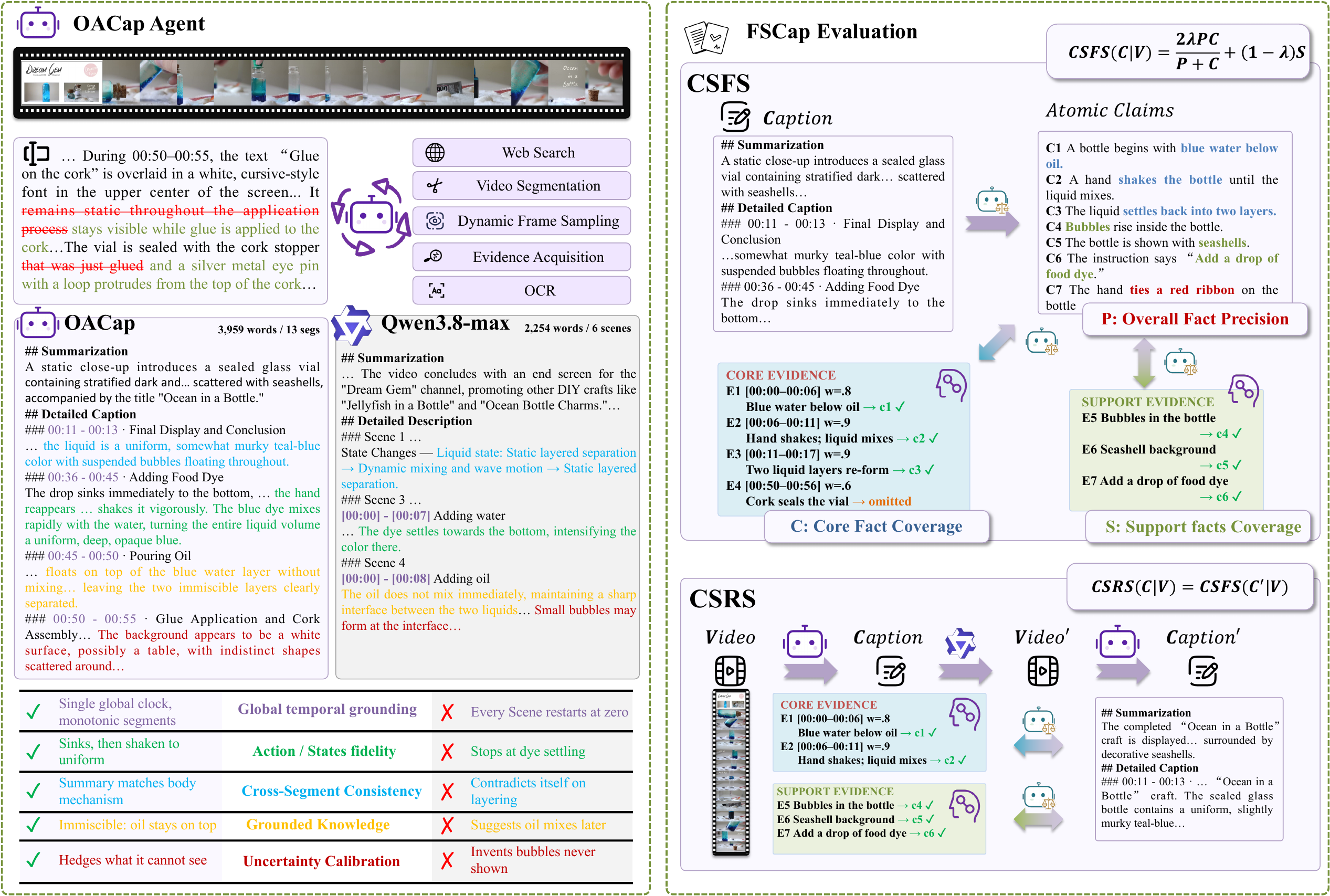}
    \caption{\textbf{OACap and FSCap-Bench at a glance.} \emph{Left:} OACap is a caption-centric video captioner that invokes tools to continually self-verify and revise its evolving caption; \cref{fig:oacap_architecture} presents the complete caption-guided observation--revision workflow. \emph{Right:} FSCap-Bench evaluates atomic claims against human-verified Core and Support facts. CSFS measures claim precision and importance-aware fact coverage, while CSRS measures visual-semantic recoverability through a fixed video--caption--video--caption round trip.}
    \label{fig:oacap_overview}
\end{figure*}

We propose \textbf{O}bservation-\textbf{A}daptive Video \textbf{Cap}tioner (OACap), illustrated on the left of \cref{fig:oacap_overview}, which treats its evolving caption as a testable account of the video. Comparing that caption with provenance-aware evidence exposes omissions, unsupported claims, conflicts, and unresolved temporal relations. OACap distinguishes \emph{revision gaps}, resolvable from existing evidence, from \emph{observation gaps}, which trigger a targeted tool call. It may crop a small object, densely sample an unclear action, apply OCR to transient text, or retrieve external evidence for entity identification. The returned evidence supports the next whole-caption revision, which in turn determines what to inspect next (\cref{fig:oacap_architecture}).

We evaluate captions as question-agnostic textual memory: a fixed text-only reader answers public video-understanding questions using captions generated without access to those questions. To diagnose quality beyond task accuracy, we also introduce \textbf{FSCap-Bench}, summarized on the right of \cref{fig:oacap_overview}. Its Core--Support Factuality Score (CSFS) combines claim precision with importance-aware coverage, while the Cross-Modal Semantic Round-Trip Score (CSRS) measures how much caption-expressed visual semantics remain recoverable through a video--caption--video--caption round trip. Controlled perturbations test sensitivity to correct additions, hallucinations, and omissions.

Our contributions are:
\begin{enumerate}[leftmargin=2em]
    \item We formulate detailed video captioning as \textbf{caption-guided active observation} and introduce OACap, which separates revision gaps resolvable from existing evidence from observation gaps requiring targeted spatial--temporal inspection.
    \item We introduce \textbf{FSCap-Bench}: 498 videos from five public benchmarks, 996 verified observation challenges, and human-verified Core and Support facts. CSFS measures factual precision and importance-aware coverage; CSRS probes recoverability through a fixed cross-modal cycle.
    \item On Video-MME, ActivityNet-QA, and VCapsBench, OACap consistently improves caption-based understanding across five matched backbones. Controlled additions, hallucinations, and omissions also show that CSFS responds correctly on at least $93\%$ of examples in every condition.
\end{enumerate}

\FloatBarrier

\section{Related Work}

\paragraph{Detailed video captioning.}
Dense video captioning localizes and describes individual events, with models such as Vid2Seq jointly predicting event boundaries and captions from narrated-video pretraining~\citep{krishna2017dense,vid2seq2023}. Recent work improves detailed captions through large-scale supervision, tool-assisted data generation, or model self-improvement: ShareGPT4Video uses differential captioning~\citep{sharegpt4video2024}, Omni-Captioner trains on detailed captions produced by a tool-assisted pipeline~\citep{omnicaptioner2025}, and VDC-Agent distills generation and self-reflection trajectories into preference data~\citep{vdcagent2025}. OACap is complementary: at inference time, it uses gaps in the evolving caption to decide what additional evidence to acquire and how to revise the whole-video description.

\paragraph{Agentic visual understanding.}
Visual agents combine language models with vision experts, executable programs, or adaptive visual operations~\citep{mmreact2023,vipergpt2023,su2025thinking,wang2024videoagent,jeoung2024adaptive,avp2025}. VideoGAIA evaluates multi-turn assistants that iteratively perceive video, invoke tools, and integrate multimodal evidence~\mbox{\citep{videogaia2026}}. Such observation is typically controlled by an external query or task state. OACap instead uses the evolving caption itself: its omissions, ambiguities, and unsupported claims determine what to inspect next.

\paragraph{Factual and semantic caption evaluation.}
Caption metrics range from reference matching to learned image--text similarity such as CLIPScore~\citep{clipscore2021}, but do not directly verify every claim in a long description. CHAIR shows that conventional scores may not penalize object hallucination~\citep{rohrbach2018chair}. FactVC and ARGUS evaluate video-caption factual errors, with ARGUS also measuring omissions~\citep{factvc2023,argus2025}; VDC, LongCaption-Bench, and VCapsBench assess detailed-caption quality or coverage~\citep{auroracap2024,longcaptioning2025,vcapsbench2026}. Correctness alone can favor short captions, whereas unweighted coverage can reward verbosity and equate central events with incidental details. CSFS instead combines claim precision with importance-aware Core and Support coverage. Cycle consistency has served as an information-preserving caption-training objective~\citep{hagiwara2019cycle}, a reconstruction-based caption metric~\citep{camscore2025}, and an image--text alignment signal~\citep{cyclereward2025}. CSRS extends this principle to detailed video captions by measuring fact recovery through a fixed video--caption--video--caption cycle.

\section{OACap: Caption-Guided Active Observation}\label{sec:pipeline}

\subsection{Agent Formulation}
Given video $V=\{v_t\}_{t=0}^{T}$, OACap produces a faithful whole-video caption without an external query. Its evolving caption supplies the observation objective: missing, unsupported, or ambiguous content becomes an evidence request. With tool set $\mathcal Z$,
\begin{equation}
C=\operatorname{Compose}(\mathcal E_N),\quad a_n=\pi(S_n),\quad
S_{n+1}=\operatorname{Update}(S_n,\operatorname{Observe}(a_n,V)),
\end{equation}
where $S_n$ is the state, $a_n$ the selected action, and $\mathcal E_N$ the verified evidence at termination.

Unlike query-driven agents, the policy cannot optimize evidence for a known answer. It must balance breadth---covering entities, events, and relations across the full video---against verification of details already proposed by the draft. The caption provides this shared interface: its claims can be checked for support, while comparison with coarse visual context exposes content that has not yet been expressed. Consequently, observation and generation are interleaved rather than executed as independent stages.

\begin{figure*}[t]
\centering
\includegraphics[width=\textwidth]{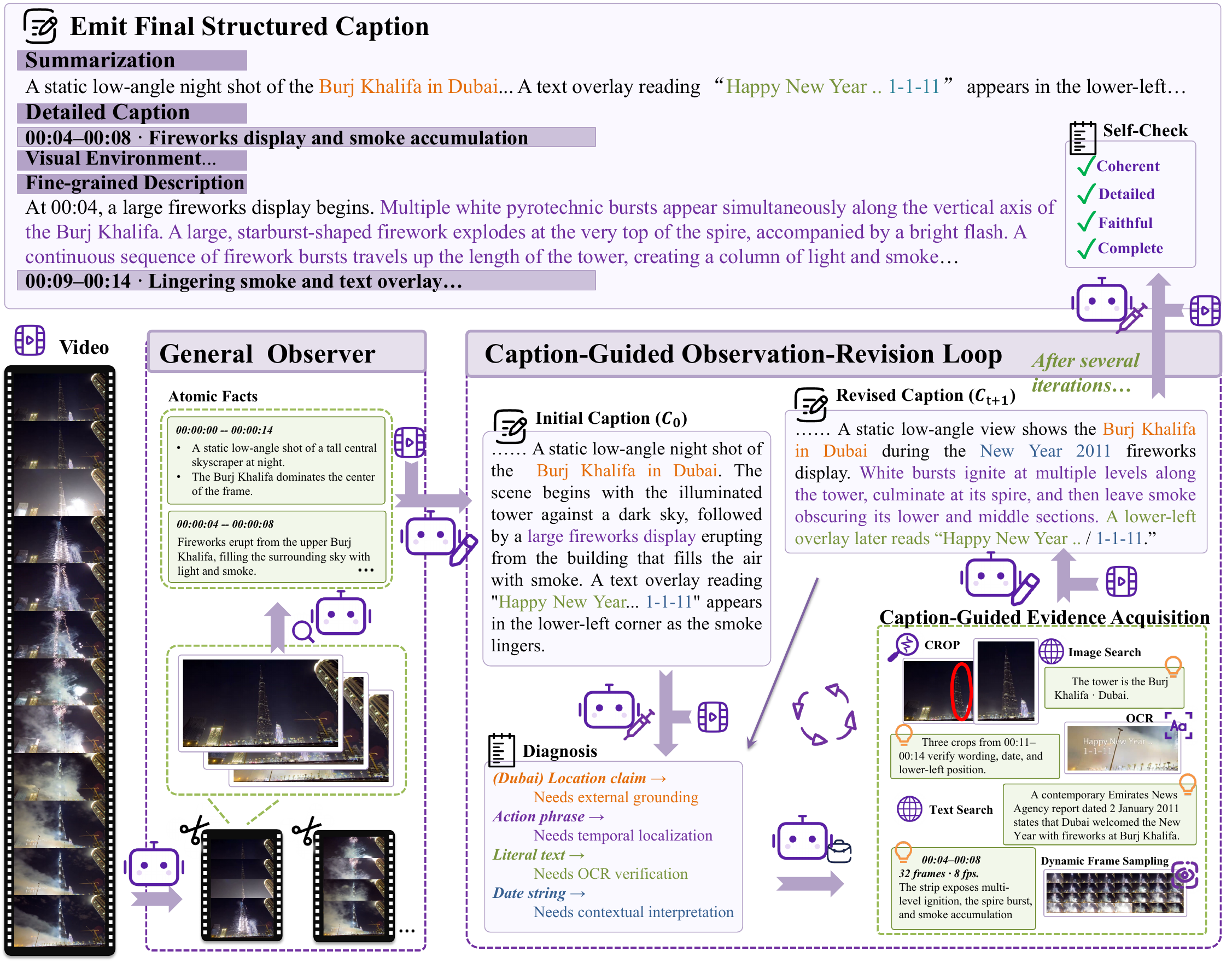}
\caption{\textbf{Caption-guided observation and revision.} Caption gaps route targeted temporal sampling, cropping, OCR, or retrieval; returned evidence supports the next revision, which exposes the next gaps.}
\label{fig:oacap_architecture}
\end{figure*}

\paragraph{Architecture overview.}
As shown in \cref{fig:oacap_architecture}, one iteration couples caption-guided acquisition with evidence-guided revision. Claim verification identifies statements that lack support, while coverage inspection uses the coarse video context to identify missing or underspecified content. The resulting gap is routed to a spatial, temporal, textual, or externally grounded observation. The returned evidence determines which claims to retain, remove, or rewrite, making the updated caption both the output of the current iteration and the control signal for the next one.

\subsection{Evidence State and Gap Diagnosis}
An inexpensive global pass yields coarse events, candidate temporal regions, and draft $C_0$. Each observation is normalized as
\begin{equation}
e_j=(f_j,\tau_j,m_j,p_j,r_j,s_j),
\end{equation}
recording its fact, interval, modality, source, confidence, and verification status (candidate, supported, conflicting, or insufficient). The state
\begin{equation}
S_n=(\mathcal E_n,C_n,\mathcal G_n,\mathcal H_n,B_n)
\end{equation}
contains evidence, the current draft, unresolved gaps, action history, and budget.

OACap treats $C_n$ as a testable hypothesis and identifies
\begin{equation}
\mathcal G_n=\mathcal G_n^{\mathrm{miss}}\cup\mathcal G_n^{\mathrm{unsupported}}
\cup\mathcal G_n^{\mathrm{conflict}}\cup\mathcal G_n^{\mathrm{temporal}}.
\end{equation}
Coverage inspection compares the draft with coarse context and accumulated evidence, allowing events absent from the text to become gaps. Each gap is expressed as a concrete evidence question. If existing evidence already resolves it, OACap edits the caption; otherwise it requests a discriminating observation.

\paragraph{Revision or re-observation.}
This distinction separates information missing from the text from information missing from the evidence state. An omitted attribute already established by an observation calls only for revision. If the available evidence cannot distinguish plausible descriptions, the agent instead seeks an observation that resolves the uncertainty: a small-object attribute calls for spatial inspection, whereas uncertainty about whether an object was inserted or removed calls for an ordered temporal sequence. Thus every tool call is tied to a possible caption change rather than a generic request for more detail.

\subsection{Observation-Adaptive Acquisition}
For a prioritized gap, OACap selects
\begin{equation}
a_n=(z_n,\tau_n,\rho_n,q_n),
\end{equation}
where $z_n$ is the tool, $\tau_n$ the temporal range, $\rho_n$ the spatial or temporal granularity, and $q_n$ the evidence question. Keyframes and crops resolve static attributes and local layout; denser sampling or clip analysis resolves trajectories and event order; OCR grounds visible text. Image retrieval or search may ground an identity or context, but external evidence is stored separately and cannot establish video actions or override contradictory visual evidence. Benchmark references are never inference inputs.

\paragraph{Illustrative trace.}
In \cref{fig:oacap_architecture}, temporal inspection corrects the claim that the on-screen text remains static by showing that ``Glue on the cork'' appears during the application step. A crop reveals the looped metal pin on the cork, while OCR recovers the separate instruction ``Add a drop of food dye.'' These observations resolve temporal, local-attribute, and textual gaps and support different edits. An inconclusive observation, by contrast, leaves its gap unresolved and cannot license a stronger caption claim.

\subsection{Evidence-Grounded Revision}
Returned observations are atomized, deduplicated, temporally aligned, and checked against $\mathcal E_n$. Agreement adds support; conflict can trigger another targeted action. OACap retains supported caption claims, corrects contradictions, and converts important unsupported claims into new gaps:
\begin{equation}
S_{n+1}=\operatorname{Update}\!\left(S_n,\operatorname{Normalize}(\operatorname{Observe}(a_n,V))\right).
\end{equation}
The loop stops when important content and temporal relations are resolved, the budget is exhausted, or expected gain is too small. The final caption is composed from verified evidence, retaining time ranges and tool traces as provenance for auditing; unresolved claims are hedged or omitted.

\section{FSCap-Bench}\label{sec:eval_protocol}
FSCap-Bench evaluates factual comprehensiveness with CSFS and cross-modal recoverability with CSRS.

\subsection{Construction}
\begin{figure*}[t]
\centering
\includegraphics[width=\textwidth]{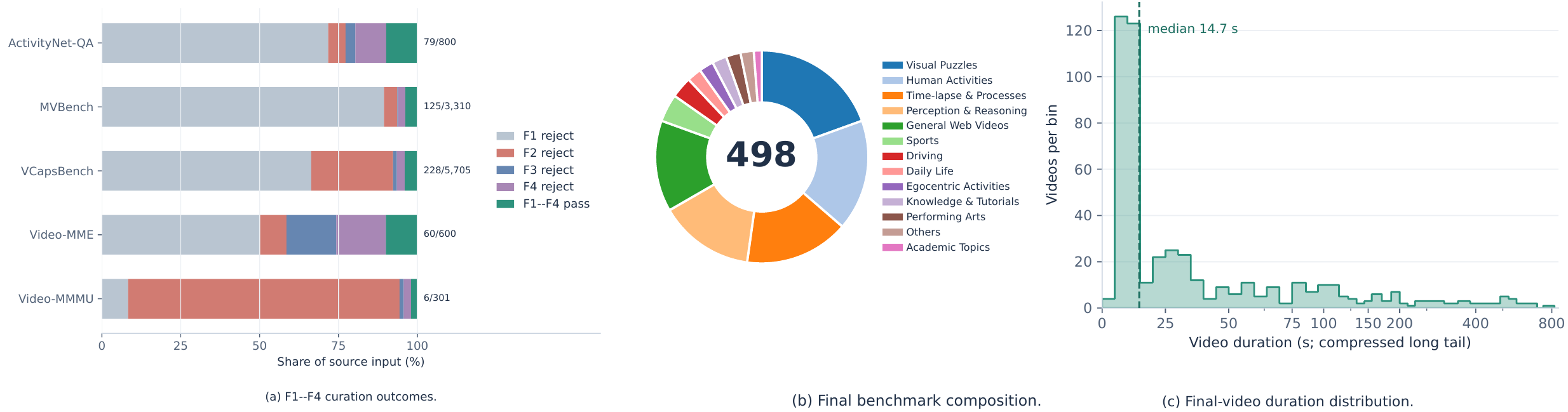}
\caption{\textbf{FSCap-Bench curation and composition.} \textbf{(a)} Terminal filtering outcomes; \textbf{(b)} category composition of the final $498$ videos; \textbf{(c)} duration distribution.}
\label{fig:selected_composition}
\end{figure*}

We curate videos from Video-MME, ActivityNet-QA, VCapsBench, Video-MMMU, and MVBench through four sequential filters: visual usability (F1), semantic density (F2), temporal coherence (F3), and caption-relevant evidence dependence (F4). F4 requires at least two verifiable challenges that benefit from targeted observation. Of $10{,}716$ candidates, $498$ pass all filters, with two verified challenges retained per video ($996$ total). \Cref{fig:selected_composition} summarizes the filtering outcomes and final distribution; per-source retention, filter criteria, and audit details are provided in \cref{sec:curation_details}.

Annotators then construct time-aligned, importance-stratified Core and Support facts for each retained video. Automatic proposals are treated only as candidates: annotators verify them against the source video, add missing facts, and adjudicate ambiguous temporal boundaries and Core/Support assignments. Full annotation and quality-control procedures are described in \cref{sec:curation_details}.

\begin{figure*}[t]
\centering
\includegraphics[width=\textwidth]{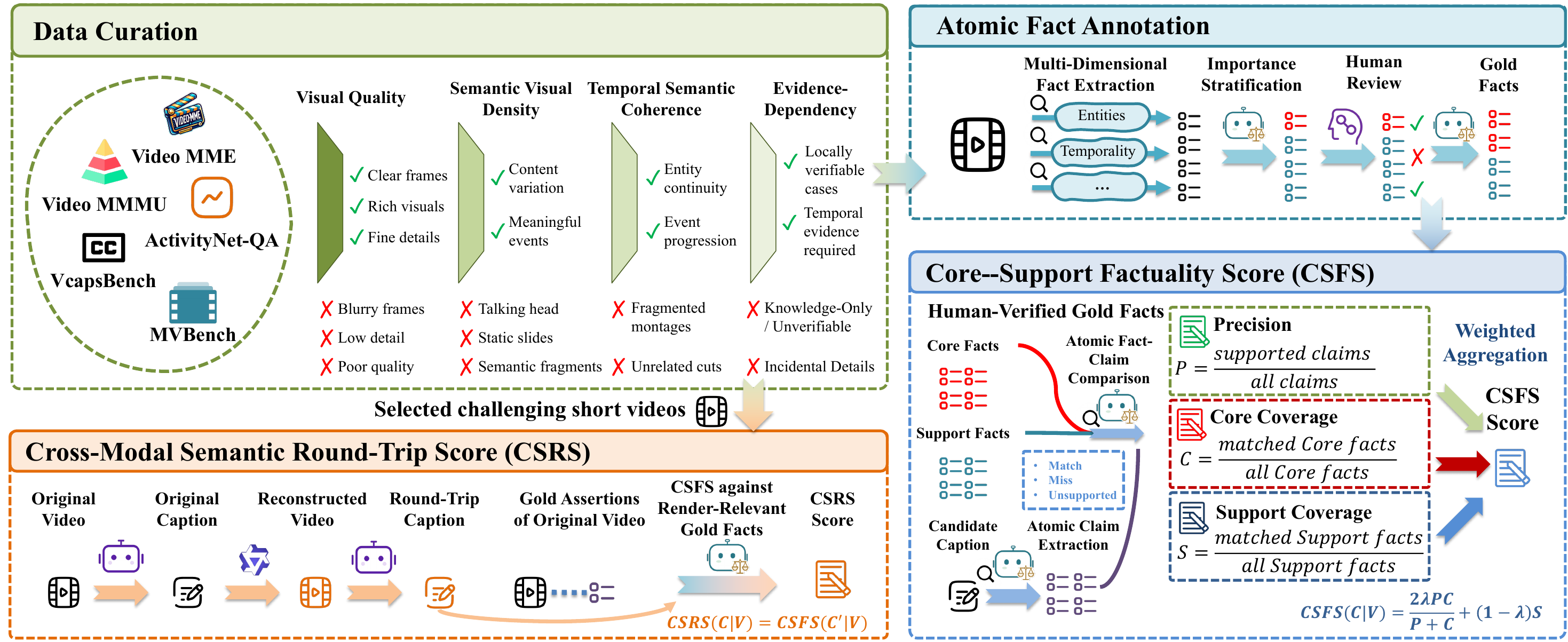}
\caption{\textbf{FSCap-Bench.} Videos are curated and annotated as Core and Support facts. CSFS measures factual precision and importance-aware coverage; CSRS matches facts before and after $V\!\to\!C\!\to\!V'\!\to\!C'$.}
\label{fig:fscap_bench_pipeline}
\end{figure*}

\subsection{Core--Support Factuality Score}\label{sec:evidence_axis}
The reference fact set is partitioned into weighted Core facts $\mathcal E_{\mathrm{core}}$ and complementary Support facts $\mathcal E_{\mathrm{sup}}$. A frozen judge atomizes caption $C$ into claims $\mathcal Q(C)$. Let $s(q)$ indicate whether claim $q$ is evidence-supported and $m(e,C)$ whether reference fact $e$ is recoverable from $C$. Then
\begin{equation}
\begin{aligned}
P&=\frac{\sum_{q\in\mathcal Q(C)}s(q)}{|\mathcal Q(C)|},\\
R_{\mathrm{core}}&=\frac{\sum_{e\in\mathcal E_{\mathrm{core}}}w_e m(e,C)}{\sum_{e\in\mathcal E_{\mathrm{core}}}w_e},\\
R_{\mathrm{sup}}&=\frac{\sum_{e\in\mathcal E_{\mathrm{sup}}}m(e,C)}{|\mathcal E_{\mathrm{sup}}|}.
\end{aligned}
\label{eq:csfs_components}
\end{equation}
Core facts describe the entities, events, and state transitions needed to preserve the video's principal semantics, with importance weights assigned before candidate captions are evaluated. Support facts capture complementary attributes and contextual detail. ``Support'' here denotes a reference category; a supported caption claim may be justified by evidence from either category. Each evaluation case contains both types, so a caption cannot obtain the aggregate score through incidental detail alone.

CSFS combines precision and Core coverage harmonically, with bounded Support credit:
\begin{equation}
\mathrm{CSFS}(C)=\lambda\frac{2P R_{\mathrm{core}}}{P+R_{\mathrm{core}}}+(1-\lambda)R_{\mathrm{sup}},\quad \lambda=0.8.
\label{eq:csfs}
\end{equation}
The judge uses the same decomposition, prompts, and matching rules for every captioner. Contradicted and unsupported claims receive zero; semantically equivalent claims are deduplicated, each reference fact is counted at most once, and reverse matches cite the supporting caption span. We report all components and macro-average per-video scores. Unsupported claims reduce $P$; omissions reduce coverage, preventing either terse or unsupported captions from scoring highly. Contradiction, unsupported-claim, temporal-order, and global-consistency rates are retained as diagnostics but are not additional factors in CSFS.

\subsection{Cross-Modal Semantic Round-Trip Score}\label{sec:roundtrip}
Given $V$, a captioner produces $C$; a text-to-video model renders $C$ as $V'$, which the same captioner describes as $C'$, forming $V\!\to\!C\!\to\!V'\!\to\!C'$. Using the same captioner at both ends avoids conflating reconstruction loss with a change in captioning model. A frozen atomizer exhaustively decomposes $C$ and $C'$ into self-contained, verifiable fact sets $A$ and $B$. A second frozen LLM jointly inspects both lists and returns semantically equivalent $(b,a)$ pairs, allowing paraphrases and differences in descriptive granularity while rejecting different objects, events, or attributes. We enforce one-to-one credit by retaining at most one returned pair for each fact in $A$ and $B$; let $M$ denote the resulting set. We also report counts $N_A=|A|$, $N_B=|B|$ and densities $D_A=100|A|/W_C$, $D_B=100|B|/W_{C'}$, where $W$ denotes word count. We compute
\begin{equation}
R=\frac{|M|}{|A|},\qquad P=\frac{|M|}{|B|},\qquad
\mathrm{CSRS}=\frac{2RP}{R+P}=\frac{2|M|}{|A|+|B|}.
\end{equation}
Here $R$ is source-fact recovery, $P$ is recovered-fact precision, Loss $=1-R$, and Drift $=1-P$ is the proportion of recovered facts absent from $C$. For empty sets, we use the explicit conventions $(R,P,\mathrm{CSRS})=(1,1,1)$ when both are empty, $(1,0,0)$ when only $A$ is empty, and $(0,1,0)$ when only $B$ is empty. We compute all scores per video before macro-averaging. CSRS measures $C$--$C'$ agreement, not correctness against $V$: an unsupported fact repeated at both ends can still match, and Drift denotes cross-modal semantic change rather than visual hallucination. CSRS thus evaluates the complete channel and complements gold-grounded CSFS. Full generation, atomization, pairing, and matching configurations are given in \cref{sec:csrs_details}.

\section{Experiments}\label{sec:experiments}

We first test whether captions serve as reusable textual memory for downstream video understanding, then use FSCap-Bench to measure factual comprehensiveness and cross-modal recoverability. All caption-based comparisons use one whole-video caption, the same text-only reader, question prompt, output budget, and abstention policy. Observation cost is reported separately because equal caption length does not imply equal inference compute. Direct-video systems are different-input references rather than caption-only competitors or strict upper bounds.

\subsection{Caption as Memory}\label{sec:caption_qa}
This experiment tests whether caption-guided observation preserves information that remains useful for questions unavailable during caption generation.
Captioners do not see downstream questions. Every single-pass baseline receives the same whole-video captioning prompt, reproduced in \cref{sec:direct_caption_prompt}. A fixed Qwen3.8-27B text-only reader receives only the caption and question, with no frames or tool traces. It must abstain when the caption is insufficient, and abstentions count as incorrect. This separation tests whether a description preserves reusable information rather than solving a known question during observation, although reader priors cannot be eliminated.

\begin{table*}[!t]
\centering
\scriptsize
\setlength{\tabcolsep}{5pt}
\caption{\textbf{Caption-as-memory results (\%).} Video-MME and ActivityNet-QA report accuracy; VCapsBench reports answer recall (AR), inconsistency (IR), and coverage (CR). The grayed \emph{oracle} row answers questions from the video itself rather than from a caption and is therefore excluded from comparisons; its overall scores also use a different input condition. Bold marks the best comparable result.}
\label{tab:caption_qa}
\begin{tabular}{lccccccc}
\toprule
\multirow{2}{*}{\textbf{Model / Method}} &
\multicolumn{3}{c}{\textbf{Video-MME}} &
\multirow{2}{*}{\textbf{ActivityNet-QA} $\uparrow$} &
\multicolumn{3}{c}{\textbf{VCapsBench}} \\
\cmidrule(lr){2-4}\cmidrule(lr){6-8}
& \textbf{Short} $\uparrow$ & \textbf{Medium} $\uparrow$ & \textbf{Long} $\uparrow$
& & \textbf{AR} $\uparrow$ & \textbf{IR} $\downarrow$ & \textbf{CR} $\uparrow$ \\
\midrule
\textcolor{gray}{Qwen3.8-27B / Direct Video QA \emph{(oracle)}} & \multicolumn{3}{c}{\textcolor{gray}{86.72 (overall)}} & \textcolor{gray}{62.10} & \multicolumn{3}{c}{\textcolor{gray}{72.94 (overall)}} \\
\midrule
Qwen3.5-A3B / Single-pass Caption  & 70.22 & 62.00 & 57.33 & 49.04 & 61.63 & 22.52 & 79.55 \\
Qwen3.5-27B / Single-pass Caption  & 78.33 & 69.89 & 65.11 & 49.22 & 66.16 & 20.84 & 83.57 \\
Qwen3.5-122B / Single-pass Caption & 76.03 & 67.89 & 64.99 & 49.07 & 64.93 & 21.48 & 82.69 \\
Qwen3.8-27B / Single-pass Caption  & 76.17 & 65.44 & 63.52 & 48.86 & 69.21 & 21.51 & 88.19 \\
Kimi-K2.6 / Single-pass Caption    & 78.01 & 71.67 & 62.33 & 49.65 & 70.35 & 19.79 & 87.71 \\
\midrule
Qwen3.5-A3B / OACap \textbf{(ours)}  & 79.39 & 73.44 & 66.95 & 49.15 & 69.78 & 19.79 & 87.00 \\
Qwen3.5-27B / OACap \textbf{(ours)}  & 78.78 & 72.79 & 66.55 & 50.29 & 70.47 & 20.50 & 88.64 \\
Qwen3.5-122B / OACap \textbf{(ours)} & 78.33 & 72.74 & \textbf{67.14} & 50.11 & 70.13 & 19.93 & 87.59 \\
Qwen3.8-27B / OACap \textbf{(ours)}  & 79.84 & 73.36 & 66.56 & \textbf{50.69} & 70.24 & 21.36 & \textbf{89.32} \\
Kimi-K2.6 / OACap \textbf{(ours)}    & \textbf{80.33} & \textbf{75.34} & 66.67 & 50.68 & \textbf{71.37} & \textbf{19.57} & 88.73 \\
\bottomrule
\end{tabular}
\end{table*}

\Cref{tab:caption_qa} compares five single-pass captioners and five model-matched OACap variants; the grayed direct-video oracle (Qwen3.8-27B answering from the video itself) is shown for reference only and is excluded from all comparisons. OACap improves every matched Video-MME split. With Qwen3.5-A3B, Short/Medium/Long accuracy rises from $70.22/62.00/57.33$ to $79.39/73.44/66.95$; with Qwen3.8-27B, it rises from $76.17/65.44/63.52$ to $79.84/73.36/66.56$. ActivityNet-QA gains are consistent ($+0.11$ to $+1.83$). On VCapsBench, every OACap variant improves AR and CR and reduces IR; OACap/Kimi-K2.6 achieves the best AR ($71.37$) and IR ($19.57$), while OACap/Qwen3.8-27B achieves the best CR ($89.32$).

The matched comparisons hold the reader and downstream protocol fixed, so the gains isolate the information made available by each captioning procedure rather than a stronger answer model. Improvements are largest on the longer Video-MME splits, where a single global pass is more likely to miss temporally separated evidence. The smaller but consistent ActivityNet-QA gains suggest that active observation still helps when questions emphasize dominant activities, while VCapsBench shows that the additional detail does not merely increase verbosity: answer and coverage rates rise as the irrelevant-information rate falls. These numbers follow our caption-as-memory protocol and are not interchangeable with official leaderboard scores that expose videos directly or use different prompting.

\subsection{CSFS Validation}\label{sec:metric_validation}
This experiment tests whether CSFS rewards supported detail, penalizes unsupported detail, and distinguishes omissions of central facts from omissions of complementary facts.
We perturb clean captions from 300 Video-MME short videos by adding approximately $20\%$ correct detail, adding approximately $20\%$ hallucinated content, deleting Core facts, or deleting Support facts, producing 1,500 captions. Three intended hallucination edits did not introduce unsupported content and are excluded from that paired condition ($n=297$); all other conditions contain 300 pairs. The expected response is to reward verified information, penalize unsupported additions and omissions, and penalize Core omissions more strongly than Support omissions.

\begin{table*}[!b]
\centering
\scriptsize
\caption{\textbf{Controlled caption perturbations} on 300 Video-MME short videos. Green arrows match the expected direction; red arrows do not.}
\label{tab:perturbation}
\setlength{\tabcolsep}{4pt}
\begin{tabular}{lccccc}
\toprule
\multirow{2}{*}{\textbf{Evaluator / configuration}} &
\multirow{2}{*}{\textbf{Clean}} &
\multicolumn{4}{c}{\textbf{Controlled perturbation}} \\
\cmidrule(lr){3-6}
& &
\textbf{+Correct} $\uparrow$ &
\textbf{+Hallucination} $\downarrow$ &
\textbf{$-$Core} $\downarrow$ &
\textbf{$-$Support} $\downarrow$ \\
\midrule
\rowcolor{LightGray}
\textcolor{gray}{\textit{Exact-provenance $D$ (Oracle)}} & \textcolor{gray}{\textit{0.9485}} & \textcolor{gray}{\textit{0.9974 $\bm{\uparrow}$}} & \textcolor{gray}{\textit{0.8696 $\bm{\downarrow}$}} & \textcolor{gray}{\textit{0.8376 $\bm{\downarrow}$}} & \textcolor{gray}{\textit{0.9168 $\bm{\downarrow}$}} \\
\midrule
PAC-S & 0.7744 & 0.7744 & 0.7748 $\textcolor{red}{\uparrow}$ & 0.7706 $\textcolor{myGreen}{\bm{\downarrow}}$ & 0.7756 $\textcolor{red}{\uparrow}$ \\
EMScore & 0.2736 & 0.2736 & 0.2736 & 0.2727 $\textcolor{myGreen}{\bm{\downarrow}}$ & 0.2743 $\textcolor{red}{\uparrow}$ \\
FactVC & 0.2261 & 0.2261 & 0.2261 & 0.2261 & 0.2266 $\textcolor{red}{\uparrow}$ \\
\midrule
BLEU-4 & 0.0268 & 0.0236 $\textcolor{red}{\downarrow}$ & 0.0226 $\textcolor{myGreen}{\bm{\downarrow}}$ & 0.0278 $\textcolor{red}{\uparrow}$ & 0.0283 $\textcolor{red}{\uparrow}$ \\
CIDEr-D & $\approx 0$ & $\approx 0$ & $\approx 0$ & $\approx 0$ & $\approx 0$ \\
BERTScore F1 & 0.7819 & 0.7816 $\textcolor{red}{\downarrow}$ & 0.7812 $\textcolor{myGreen}{\bm{\downarrow}}$ & 0.7816 $\textcolor{myGreen}{\bm{\downarrow}}$ & 0.7828 $\textcolor{red}{\uparrow}$ \\
\midrule
ARGUS-Qwen & 0.5028 & 0.4978 $\textcolor{red}{\downarrow}$ & 0.5115 $\textcolor{red}{\uparrow}$ & 0.4607 $\textcolor{myGreen}{\bm{\downarrow}}$ & 0.4982 $\textcolor{myGreen}{\bm{\downarrow}}$ \\
VDCScore-Qwen & 4.9457 & 4.9668 $\textcolor{myGreen}{\bm{\uparrow}}$ & 4.9460 $\textcolor{red}{\uparrow}$ & 4.6820 $\textcolor{myGreen}{\bm{\downarrow}}$ & 4.7410 $\textcolor{myGreen}{\bm{\downarrow}}$ \\
CapST-Qwen & 0.7561 & 0.7398 $\textcolor{red}{\downarrow}$ & 0.7537 $\textcolor{myGreen}{\bm{\downarrow}}$ & 0.7386 $\textcolor{myGreen}{\bm{\downarrow}}$ & 0.7608 $\textcolor{red}{\uparrow}$ \\
Direct Qwen Scene Judge & 0.3539 & 0.3435 $\textcolor{red}{\downarrow}$ & 0.2788 $\textcolor{myGreen}{\bm{\downarrow}}$ & 0.3600 $\textcolor{red}{\uparrow}$ & 0.3583 $\textcolor{red}{\uparrow}$ \\
\midrule
\textbf{CSFS (ours)} & \textbf{0.9291} & \textbf{0.9694} $\textcolor{myGreen}{\bm{\uparrow}}$ & \textbf{0.8752} $\textcolor{myGreen}{\bm{\downarrow}}$ & \textbf{0.8836} $\textcolor{myGreen}{\bm{\downarrow}}$ & \textbf{0.9039} $\textcolor{myGreen}{\bm{\downarrow}}$ \\
\bottomrule
\end{tabular}
\end{table*}

CSFS follows all intended directions (\cref{tab:perturbation}): correct detail raises $0.9291\to0.9694$, hallucination lowers it to $0.8752$, and Core/Support deletion lowers it to $0.8836/0.9039$. Its components respond selectively: correct additions raise Support coverage from $0.6677$ to $0.8734$; hallucinations reduce precision from $0.9987$ to $0.8853$; Core deletion reduces weighted Core coverage from $0.9919$ to $0.8979$; and Support deletion reduces Support coverage to $0.5384$. At the video level, directional accuracy is $97.3\%$, $98.3\%$, $93.0\%$, and $93.3\%$, respectively---the only automatic method exceeding $93\%$ in all four conditions.

Baselines fail for complementary reasons. The tested PAC-S, EMScore, and FactVC configurations truncate long captions to 77 tokens; similarity metrics weakly or inconsistently respond to fine-grained edits; and Qwen adaptations do not reliably distinguish correct additions from hallucinations. These results validate directional sensitivity, but not human agreement or invariance to alternative atomic decompositions.

\subsection{CSRS Validation}
This experiment tests two complementary claims: whether OACap improves factual comprehensiveness under matched captioning backbones, and how much caption-expressed information survives a fixed cross-modal reconstruction channel.
\Cref{tab:csfs_results} compares direct captioning with OACap under matched backbones on FSCap-Bench, while \cref{tab:csrs_results} evaluates recoverability through the fixed reconstruction channel. OACap improves CSFS for all three backbones, with gains of $1.94$, $5.20$, and $20.75$ points for Qwen3.5-27B, Qwen3.5-122B, and Qwen3.8-27B, respectively.

For CSRS, we evaluate six configurations on the same 33 FSCap-Bench videos sourced from Video-MME, all shorter than two minutes. Qwen3.8-27B or Qwen3.8-Max produces $C$ and is reused to caption $V'$. LTX-2.5-Diffusers renders both 10- and 60-second videos, while Wan2.2-T2V-A14B renders 10-second videos, giving $2\times2$ generator--captioner combinations at 10 seconds and two additional LTX configurations at 60 seconds. A frozen atomizer extracts source and recovered facts, and a separate frozen judge returns one-to-one semantic matches; \cref{sec:csrs_details} specifies the complete protocol.

\begin{table}[!htbp]
\centering
\scriptsize
\caption{\textbf{Factual comprehensiveness on FSCap-Bench.} CSFS is reported on a $0$--$100$ scale. Direct Caption uses a single whole-video caption generated by each backbone, while OACap Pipeline applies our caption-guided observation and revision framework with the same backbone. $\Delta$ is the absolute improvement in CSFS points and is computed from unrounded scores.}
\label{tab:csfs_results}
\begin{tabular}{lccc}
\toprule
\textbf{Backbone} & \textbf{Direct Caption} & \textbf{OACap Pipeline} & $\boldsymbol{\Delta}\uparrow$ \\
\midrule
Qwen3.5-27B  & 65.87 & 67.81 & +1.94 \\
Qwen3.5-122B & 63.38 & 68.58 & +5.20 \\
Qwen3.8-27B  & \textbf{66.50} & \textbf{87.24} & \textbf{+20.75} \\
\bottomrule
\end{tabular}
\end{table}

\begin{table*}[!htbp]
\centering
\scriptsize
\setlength{\tabcolsep}{7.0pt}
\caption{\textbf{Cross-modal round-trip results} on 33 FSCap-Bench videos sourced from Video-MME, all shorter than two minutes (per-video macro average). $R$: source-fact recovery; $P$: recovered-fact precision; CSRS: fact-level $F_1$; Loss$=1-R$; Drift$=1-P$. LTX is evaluated at 10 and 60 seconds; Wan is evaluated at 10 seconds.}
\label{tab:csrs_results}
\begin{tabular}{lrrrrr}
\toprule
\textbf{T2V / Duration / Captioner} & $R\uparrow$ & $P\uparrow$ & \textbf{CSRS}$\uparrow$ & Loss$\downarrow$ & Drift$\downarrow$ \\
\midrule
LTX-2.5 / 10s / Qwen3.8-27B & 0.206 & 0.352 & 0.245 & 0.794 & 0.648 \\
LTX-2.5 / 10s / Qwen3.8-Max & 0.230 & 0.257 & 0.231 & 0.770 & 0.743 \\
Wan2.2 / 10s / Qwen3.8-27B & 0.168 & 0.296 & 0.206 & 0.832 & 0.704 \\
Wan2.2 / 10s / Qwen3.8-Max & 0.187 & 0.241 & 0.200 & 0.813 & 0.759 \\
LTX-2.5 / 60s / Qwen3.8-27B & 0.213 & \textbf{0.362} & 0.254 & 0.787 & \textbf{0.638} \\
LTX-2.5 / 60s / Qwen3.8-Max & \textbf{0.291} & 0.327 & \textbf{0.291} & \textbf{0.709} & 0.673 \\
\bottomrule
\end{tabular}
\end{table*}

At 10 seconds, LTX preserves more caption-expressed information than Wan for both captioners: CSRS improves from $0.206$ to $0.245$ with Qwen3.8-27B and from $0.200$ to $0.231$ with Qwen3.8-Max. Extending LTX to 60 seconds raises CSRS to $0.254$ and $0.291$, respectively. The captioners expose a recall--precision trade-off: Qwen3.8-Max generally recovers more source facts, whereas Qwen3.8-27B yields higher precision and lower drift. LTX-60s/Qwen3.8-Max achieves the best overall CSRS ($0.291$) and recovery ($0.291$); LTX-60s/Qwen3.8-27B achieves the best precision ($0.362$) and lowest drift ($0.638$). Absolute information loss remains high ($70.9\%$--$83.2\%$), reflecting the complete caption--generation--caption channel rather than the T2V model alone.

\section{Limitations}

OACap trades additional observation and model calls for caption quality. The present comparisons hold the downstream reader fixed but do not establish a quality--cost Pareto frontier against equal-compute uniform sampling or alternative multi-stage policies. The routing and stopping decisions also inherit the perceptual and calibration errors of the models that diagnose caption gaps and interpret tool outputs.

The evaluations have distinct scope limits. Caption-as-memory scores depend on reader priors and question coverage. CSFS depends on annotation completeness, importance assignments, atomic decomposition, and judge reliability; an omitted reference can make a correct claim appear unsupported. Because FSCap-Bench draws from datasets also used in the downstream evaluation, claims of independent generalization require a video-level overlap audit. CSRS combines errors from the generator, re-captioner, atomizer, and matcher, and the current 33-video subset covers only generation-compatible visual content. We therefore interpret downstream utility, CSFS, and CSRS jointly rather than as interchangeable estimates of caption quality.

\section{Conclusion}

OACap treats a detailed caption as both the output and the control state of video understanding. By separating gaps that can be repaired from existing evidence from gaps that require new observation, it turns caption deficiencies into targeted tool calls and evidence-grounded revisions. Across matched backbones, the resulting captions preserve more information for downstream question answering and achieve higher factual comprehensiveness on FSCap-Bench. CSFS and CSRS provide complementary views of this improvement: the former measures supported claims and importance-aware coverage against video-grounded facts, whereas the latter measures information recoverability through a fixed cross-modal channel. Together, the method and evaluation framework support a broader view of captioning as active evidence acquisition rather than one-pass description generation.

\clearpage
\bibliographystyle{antgroup}

\bibliography{ref/reference}

\begin{thebibliography}{23}
\providecommand{\natexlab}[1]{#1}
\providecommand{\url}[1]{\texttt{#1}}
\expandafter\ifx\csname urlstyle\endcsname\relax
  \providecommand{\doi}[1]{doi: #1}\else
  \providecommand{\doi}{doi: \begingroup \urlstyle{rm}\Url}\fi

\bibitem[Bahng et~al.(2025)Bahng, Chan, Durand, and Isola]{cyclereward2025}
Hyojin Bahng, Caroline Chan, Fredo Durand, and Phillip Isola.
\newblock {Cycle Consistency as Reward: Learning Image-Text Alignment without
  Human Preferences}.
\newblock In \emph{Proceedings of the IEEE/CVF International Conference on
  Computer Vision (ICCV)}, pp.\  22934--22946, 2025.

\bibitem[Chai et~al.(2025)Chai, Song, Du, Meng, Madhavan, Bar-Tal, Hwang, Xie,
  and Manning]{auroracap2024}
Wenhao Chai, Enxin Song, Yilun Du, Chenlin Meng, Vashisht Madhavan, Omer
  Bar-Tal, Jenq-Neng Hwang, Saining Xie, and Christopher~D. Manning.
\newblock {AuroraCap: Efficient, Performant Video Detailed Captioning and a New
  Benchmark}.
\newblock In \emph{The Thirteenth International Conference on Learning
  Representations (ICLR)}, 2025.
\newblock OpenReview ID: tTDUrseRRU.

\bibitem[Chen et~al.(2024)Chen, Wei, Li, Dong, Zhang, Zang, Chen, Duan, Lin,
  Tang, et~al.]{sharegpt4video2024}
Lin Chen, Xilin Wei, Jinsong Li, Xiaoyi Dong, Pan Zhang, Yuhang Zang, Zehui
  Chen, Haodong Duan, Bin Lin, Zhenyu Tang, et~al.
\newblock {ShareGPT4Video: Improving Video Understanding and Generation with
  Better Captions}.
\newblock \emph{arXiv preprint arXiv:2406.04325}, 2024.

\bibitem[Cui et~al.(2025)Cui, Bai, Wang, Chen, Xu, Luo, Zhang, and
  Shi]{camscore2025}
Tianyu Cui, Jinbin Bai, Guo-Hua Wang, Qing-Guo Chen, Zhao Xu, Weihua Luo, Kaifu
  Zhang, and Ye~Shi.
\newblock {Evaluating Image Caption via Cycle-consistent Text-to-Image
  Generation}.
\newblock \emph{arXiv preprint arXiv:2501.03567}, 2025.

\bibitem[Hagiwara et~al.(2019)Hagiwara, Mukuta, and Harada]{hagiwara2019cycle}
Keisuke Hagiwara, Yusuke Mukuta, and Tatsuya Harada.
\newblock {End-to-End Learning Using Cycle Consistency for Image-to-Caption
  Transformations}.
\newblock \emph{arXiv preprint arXiv:1903.10118}, 2019.

\bibitem[Hessel et~al.(2021)Hessel, Holtzman, Forbes, Le~Bras, and
  Choi]{clipscore2021}
Jack Hessel, Ari Holtzman, Maxwell Forbes, Ronan Le~Bras, and Yejin Choi.
\newblock {CLIPScore: A Reference-free Evaluation Metric for Image Captioning}.
\newblock In \emph{Proceedings of the 2021 Conference on Empirical Methods in
  Natural Language Processing}, pp.\  7514--7528, 2021.
\newblock \doi{10.18653/v1/2021.emnlp-main.595}.

\bibitem[Jeoung et~al.(2024)Jeoung, Huybrechts, Ganesh, Galstyan, and
  Bodapati]{jeoung2024adaptive}
Sullam Jeoung, Goeric Huybrechts, Bhavana Ganesh, Aram Galstyan, and Sravan
  Bodapati.
\newblock {Adaptive Video Understanding Agent: Enhancing Efficiency with
  Dynamic Frame Sampling and Feedback-Driven Reasoning}.
\newblock NeurIPS 2024 Workshop on Multimodal Algorithmic Reasoning (MAR),
  2024.
\newblock Oral presentation; non-archival workshop.

\bibitem[Krishna et~al.(2017)Krishna, Hata, Ren, Fei-Fei, and
  Niebles]{krishna2017dense}
Ranjay Krishna, Kenji Hata, Frederic Ren, Li~Fei-Fei, and Juan~Carlos Niebles.
\newblock {Dense-Captioning Events in Videos}.
\newblock In \emph{Proceedings of the IEEE International Conference on Computer
  Vision (ICCV)}, 2017.

\bibitem[Liu \& Wan(2023)Liu and Wan]{factvc2023}
Hui Liu and Xiaojun Wan.
\newblock Models see hallucinations: Evaluating the factuality in video
  captioning.
\newblock In \emph{Proceedings of the 2023 Conference on Empirical Methods in
  Natural Language Processing}, pp.\  11807--11823, Singapore, December 2023.
  Association for Computational Linguistics.
\newblock \doi{10.18653/v1/2023.emnlp-main.723}.

\bibitem[Ma et~al.(2026)Ma, Xu, Xing, Chu, Wang, He, Xu, Heng, Yu, Lin, Chng,
  and Chen]{omnicaptioner2025}
Ziyang Ma, Ruiyang Xu, Zhenghao Xing, Yunfei Chu, Yuxuan Wang, Jinzheng He, Jin
  Xu, Pheng-Ann Heng, Kai Yu, Junyang Lin, Eng-Siong Chng, and Xie Chen.
\newblock {Omni-Captioner: Data Pipeline, Models, and Benchmark for Omni
  Detailed Perception}.
\newblock In \emph{The Fourteenth International Conference on Learning
  Representations (ICLR)}, 2026.
\newblock OpenReview ID: Z091XLyVkJ.

\bibitem[Rawal et~al.(2025)Rawal, Shirkavand, Huang, Somepalli, and
  Goldstein]{argus2025}
Ruchit Rawal, Reza Shirkavand, Heng Huang, Gowthami Somepalli, and Tom
  Goldstein.
\newblock {ARGUS: Hallucination and Omission Evaluation in Video-LLMs}.
\newblock In \emph{Proceedings of the IEEE/CVF International Conference on
  Computer Vision (ICCV)}, pp.\  20280--20290. IEEE, October 2025.
\newblock \doi{10.1109/ICCV51701.2025.01886}.

\bibitem[Rohrbach et~al.(2018)Rohrbach, Hendricks, Burns, Darrell, and
  Saenko]{rohrbach2018chair}
Anna Rohrbach, Lisa~Anne Hendricks, Kaylee Burns, Trevor Darrell, and Kate
  Saenko.
\newblock Object hallucination in image captioning.
\newblock In \emph{Proceedings of the 2018 Conference on Empirical Methods in
  Natural Language Processing}, pp.\  4035--4045. Association for Computational
  Linguistics, 2018.
\newblock \doi{10.18653/v1/D18-1437}.

\bibitem[Su et~al.(2025)Su, Xia, Guo, Liu, Ma, Qu, Liu, Li, Zeng, Yang, Li,
  Cheng, Ji, He, and Fung]{su2025thinking}
Zhaochen Su, Peng Xia, Hangyu Guo, Zhenhua Liu, Yan Ma, Xiaoye Qu, Jiaqi Liu,
  Yanshu Li, Kaide Zeng, Zhengyuan Yang, Linjie Li, Yu~Cheng, Heng Ji, Junxian
  He, and Yi~R. Fung.
\newblock {Thinking with Images for Multimodal Reasoning: Foundations, Methods,
  and Future Frontiers}.
\newblock \emph{arXiv preprint arXiv:2506.23918}, 2025.

\bibitem[Sur{\'\i}s et~al.(2023)Sur{\'\i}s, Menon, and Vondrick]{vipergpt2023}
D{\'\i}dac Sur{\'\i}s, Sachit Menon, and Carl Vondrick.
\newblock {ViperGPT: Visual Inference via Python Execution for Reasoning}.
\newblock In \emph{Proceedings of the IEEE/CVF International Conference on
  Computer Vision (ICCV)}, pp.\  11888--11898, 2023.
\newblock \doi{10.1109/ICCV51070.2023.01092}.

\bibitem[Tang et~al.(2025)Tang, Li, Yang, Zhuang, Sun, Li, Ma, and
  Zhang]{videosalmonn2025}
Changli Tang, Yixuan Li, Yudong Yang, Jimin Zhuang, Guangzhi Sun, Wei Li, Zejun
  Ma, and Chao Zhang.
\newblock {video-SALMONN 2: Captioning-Enhanced Audio-Visual Large Language
  Models}.
\newblock \emph{arXiv preprint arXiv:2506.15220}, 2025.

\bibitem[Wang et~al.(2026)Wang, Gao, He, Han, Li, Dong, Ma, and
  Gong]{vdcagent2025}
Qiang Wang, Xinyuan Gao, Yuhang He, Jizhou Han, Jiangyang Li, Songlin Dong,
  Zhiheng Ma, and Yihong Gong.
\newblock {VDC-Agent: When Video Detailed Captioners Evolve Themselves via
  Agentic Self-Reflection}.
\newblock In \emph{European Conference on Computer Vision (ECCV)}, pp.\
  515--535, 2026.

\bibitem[Wang et~al.(2024)Wang, Zhang, Zohar, and
  Yeung-Levy]{wang2024videoagent}
Xiaohan Wang, Yuhui Zhang, Orr Zohar, and Serena Yeung-Levy.
\newblock {VideoAgent: Long-form Video Understanding with Large Language Model
  as Agent}.
\newblock \emph{arXiv preprint arXiv:2403.10517}, 2024.

\bibitem[Wang et~al.(2025)Wang, Zhou, Wang, Li, Xiong, Savarese, Bansal, Ryoo,
  and Niebles]{avp2025}
Ziyang Wang, Honglu Zhou, Shijie Wang, Junnan Li, Caiming Xiong, Silvio
  Savarese, Mohit Bansal, Michael~S. Ryoo, and Juan~Carlos Niebles.
\newblock {Active Video Perception: Iterative Evidence Seeking for Agentic Long
  Video Understanding}.
\newblock \emph{arXiv preprint arXiv:2512.05774}, 2025.

\bibitem[Wei et~al.(2025)Wei, Tan, Hu, Chen, and Chen]{longcaptioning2025}
Hongchen Wei, Zhihong Tan, Yaosi Hu, Chang~Wen Chen, and Zhenzhong Chen.
\newblock {LongCaptioning: Unlocking the Power of Long Video Caption Generation
  in Large Multimodal Models}.
\newblock \emph{arXiv preprint arXiv:2502.15393}, 2025.

\bibitem[Yang et~al.(2023{\natexlab{a}})Yang, Nagrani, Seo, Miech, Pont-Tuset,
  Laptev, Sivic, and Schmid]{vid2seq2023}
Antoine Yang, Arsha Nagrani, Paul~Hongsuck Seo, Antoine Miech, Jordi
  Pont-Tuset, Ivan Laptev, Josef Sivic, and Cordelia Schmid.
\newblock {Vid2Seq: Large-Scale Pretraining of a Visual Language Model for
  Dense Video Captioning}.
\newblock In \emph{Proceedings of the IEEE/CVF Conference on Computer Vision
  and Pattern Recognition (CVPR)}, pp.\  10714--10726, 2023{\natexlab{a}}.

\bibitem[Yang et~al.(2023{\natexlab{b}})Yang, Li, Wang, Lin, Azarnasab, Ahmed,
  Liu, Liu, Zeng, and Wang]{mmreact2023}
Zhengyuan Yang, Linjie Li, Jianfeng Wang, Kevin Lin, Ehsan Azarnasab, Faisal
  Ahmed, Zicheng Liu, Ce~Liu, Michael Zeng, and Lijuan Wang.
\newblock {MM-ReAct: Prompting ChatGPT for Multimodal Reasoning and Action}.
\newblock \emph{arXiv preprint arXiv:2303.11381}, 2023{\natexlab{b}}.

\bibitem[Zhang et~al.(2026{\natexlab{a}})Zhang, Yao, Wu, Wu, Lian, Geng, Chen,
  Yuan, and Heng]{videogaia2026}
Fan Zhang, Guangming Yao, Jinyang Wu, Hao Wu, Zheng Lian, Xinyu Geng, Jingdong
  Chen, Yi~Yuan, and Pheng-Ann Heng.
\newblock {VideoGAIA: A Benchmark for General AI Assistants on Agentic Video
  Understanding}.
\newblock \emph{arXiv preprint arXiv:2608.14718}, 2026{\natexlab{a}}.

\bibitem[Zhang et~al.(2026{\natexlab{b}})Zhang, Wang, Huang, Li, Zhu, and
  Yin]{vcapsbench2026}
Shi-Xue Zhang, Hongfa Wang, Duojun Huang, Xin Li, Xiaobin Zhu, and Xu-Cheng
  Yin.
\newblock {VCapsBench: A Large-scale Fine-grained Benchmark for Video Caption
  Quality Evaluation}.
\newblock In \emph{Proceedings of the AAAI Conference on Artificial
  Intelligence}, pp.\  12726--12734, 2026{\natexlab{b}}.
\newblock \doi{10.1609/aaai.v40i15.38269}.

\end{thebibliography}

\clearpage
\appendix
\section{Curation Protocol and Coverage}\label{sec:curation_details}

\subsection{Benchmark Composition}

\paragraph{Data provenance.}
The curation manifest contains $10{,}716$ unique video IDs with item-level media metadata, stage reachability, measured gate outputs, terminal-stage decisions, and F4 verification records. Durations span $2.08$--$2{,}029.63$ seconds. All counts in \cref{fig:selected_composition} are computed from this manifest rather than transcribed aggregates. Terminal stages are mutually exclusive, while diagnostic counterfactual readings from unreached stages are retained separately and never counted as funnel outcomes.

Overall, $4.6\%$ of candidates survive all four filters. Retention is $60/600$ ($10.0\%$) for Video-MME, $79/800$ ($9.9\%$) for ActivityNet-QA, $228/5{,}705$ ($4.0\%$) for VCapsBench, $125/3{,}310$ ($3.8\%$) for MVBench, and $6/301$ ($2.0\%$) for Video-MMMU. After merging source-specific labels, the largest groups are Visual Puzzles, Human Activities, Time-lapse \& Processes, Perception \& Reasoning, and General Web Videos. The set is intentionally short-video-skewed (median $14.7$ seconds), but retains a long tail: $433/498$ videos are under two minutes. These statistics characterize the selected evaluation distribution rather than the prevalence of observation challenges in the source datasets.

\subsection{Four-Stage Filtering}

\paragraph{F1: visual usability.}
Twelve uniformly spaced midpoint frames provide grayscale sharpness, high-frequency detail, brightness, and invalid-frame statistics. Brightness-normalized Laplacian variance reduces dark-scene bias. F1 rejects a video if quality is below $0.60$, blurred-frame fraction exceeds $0.30$, minimum resolution is below $480$p, black-frame fraction exceeds $0.15$, or invalid-frame fraction exceeds $0.25$. The quality score combines sharpness, detail, non-blurred fraction, bitrate per pixel per frame, and resolution. F1 is evaluated locally without VLM calls and rejects $7{,}645$ videos, leaving $3{,}071$.

\paragraph{F2: semantic visual content.}
Bounded local windows first remove hard static or image-like cases, rejecting $304$ videos. For the remaining candidates, a VLM reads timestamped frames and must return keep$=1$, density$\geq20$, events$\geq4$, variety$\geq2$, and no banned image-content type. The VLM gates reject another $1{,}668$ videos, leaving $1{,}099$. Less restrictive static, dialog, and content-type indicators remain diagnostic only. These exclusions define the benchmark scope and can remove visually informative instructional or screen content; low retention alone does not validate rejection precision.

\paragraph{F3: temporal coherence.}
A VLM reads timestamped frames and distinguishes trackable processes from unrelated montage. F3 requires event density$\geq45$, montage noise$\leq0.5$, causality$\geq3$, and entity continuity$\geq2$. Shifted-sampling probes are retained when present, and passing the measured funnel decision determines eligibility. F3 rejects $202$ videos and sends $897$ to F4. Strong-causality gating may exclude coherent events without an explicit causal chain, motivating separate rejection-set auditing.

\paragraph{F4 audit record.}
Each proposed challenge records a fact, evidence anchor, caption impact, required observation, and an optional plausible alternative. A separate verifier checks the indicated region or segment. Irrelevant decoration, unsupported intent, and external-knowledge-only difficulty do not qualify. The release gate requires at least two verified challenges per video and rejects knowledge-only cases; two verified challenges are retained for each passing video. Among the $897$ candidates reaching F4, $399$ are rejected and $498$ pass, yielding $996$ valid challenge records. State changes are the most frequent challenge type ($364$ videos), followed by local attributes ($228$) and on-screen text ($163$); types may overlap within a video. Processing failures remain unresolved rather than being counted as easy cases.

\subsection{Fact Annotation and Quality Control}

Annotators create time-aligned atomic facts spanning attributes, spatial relations, visible text, actions, transient motion, state changes and order, long-range relations, and audio-visual evidence. Automatic multi-tool proposals are treated only as candidates: annotators verify them against the source video and add missing important facts. Facts are partitioned into importance-weighted Core facts and complementary Support facts. Ambiguous temporal boundaries and Core/Support assignments are adjudicated. Externally grounded knowledge requires a visible anchor and never substitutes for evidence that a video event occurred.

\section{Evaluation Prompts and Round-Trip Configuration}\label{sec:evaluation_details}

\subsection{Single-Pass Direct-Caption Prompt}\label{sec:direct_caption_prompt}
All single-pass captioners in \cref{tab:caption_qa} receive the same prompt below. The prompt is question-agnostic and requests one temporally structured whole-video description.

\begin{tcblisting}{
  listing only,
  breakable,
  colback=black!2,
  colframe=black!35,
  boxrule=0.4pt,
  arc=1pt,
  left=5pt,
  right=5pt,
  top=4pt,
  bottom=4pt,
  listing options={
    basicstyle=\small\rmfamily,
    columns=fullflexible,
    keepspaces=true,
    breaklines=true,
    showstringspaces=false
  }
}
Watch the entire supplied video and write a faithful, coherent, and
appropriately detailed English caption. The video is the only
observational source. Describe its beginning, intermediate phases,
and ending in chronological order.

Use this Markdown structure:

## Summary

A concise account of the main subjects, events, and outcome actually
visible.

## Detailed Caption

### MM:SS - MM:SS: Descriptive phase title

Connected prose describing this meaningful phase of the video.
Repeat phase blocks as needed to cover the full video.

Timestamp requirements:

- Use absolute video times rounded to whole seconds.
- Use HH:MM:SS for videos longer than one hour.
- Keep blocks chronological, within the actual video duration, and
  non-overlapping.
- Every block must have a positive duration: the end time must be
  later than the start time.
- Split phases when the setting, principal action, participants,
  topic, or object state changes meaningfully.
- Do not create a separate block for every camera cut when the same
  event continues.
- Brief events lasting less than one second may be described inside
  the surrounding phase. Do not create zero-duration blocks or
  falsely extend a brief event.
- Treat timestamps as useful approximate boundaries rather than
  fabricated frame-level precision.

Coverage and writing requirements:

- Cover the beginning, intermediate developments, and ending.
- Give greater detail to information-rich or rapidly changing
  portions.
- Preserve distinct stages even when they occur in the same setting.
- Mention repeated actions only when the repetition or count is
  meaningful.
- Avoid repeating the same description across adjacent blocks.
- Do not turn the caption into a shot-by-shot camera log unless
  camera movement affects understanding.
- Describe only what can be supported by the supplied video. Do not
  infer unseen actions or guess unclear details.
- Do not include analysis, speculation, external knowledge, hidden
  reasoning, QA answers, or comments about the captioning process.
- Do not use tables, JSON, confidence scores, or additional top-level
  sections.
- Return only the requested Markdown caption.
\end{tcblisting}

\subsection{CSRS Configuration and Fact Extraction}\label{sec:csrs_details}
\paragraph{Configurations and pairing.}
All configurations share the same 33 FSCap-Bench videos sourced from Video-MME; every selected video is shorter than two minutes. At 10 seconds, we cross LTX-2.5-Diffusers and Wan2.2-T2V-A14B with Qwen3.8-27B and Qwen3.8-Max, producing four configurations. Because the evaluated Wan pipeline supports only the short setting, the 60-second evaluation contains the two LTX--captioner combinations, for six configurations and 198 source--recovered caption pairs in total. For each pair, a captioner produces source caption $C$ from the original video, the T2V model renders $C$ into $V'$, and the \emph{same} captioner produces $C'$ from $V'$. The captioners use temperature $0.2$, a 32,768-token output limit, disabled explicit thinking, and a common duration-aware frame prompt. Records are joined by video ID and verified through the stored SHA-256 hash of $C$, preventing accidental cross-captioner or cross-configuration pairing.

\paragraph{Atomic-fact extraction.}
Qwen3.8-Max serves as a fixed atomizer for both $C$ and $C'$, with temperature $0$, a 16,384-token output limit, disabled explicit thinking, a 300-second timeout, and at most three attempts separated by two seconds. The prompt requests only a JSON object containing an exhaustive Chinese-language fact list. Every fact must be self-contained, concrete, observable, non-speculative, and limited to one assertion. Composite descriptions are split so that people and clothing, object attributes, colors, spatial relations, actions, temporal order, visible text, camera motion, lighting, environment, transitions, and audio cues can be matched independently. Quoted on-screen text is normalized to Chinese book-title brackets to reduce malformed JSON. We parse, in order, the complete response, a fenced JSON block, or the substring from the first opening to the last closing brace; a final regular-expression fallback recovers quoted items from the JSON fact array. Each distinct caption is processed once and cached by its SHA-256 hash.

\paragraph{Semantic matching.}
Kimi-K3-SPOT serves as the fixed matcher, with temperature $0$, a 4,096-token output limit, disabled explicit thinking, a 180-second timeout, and the same three-attempt retry policy. It receives the complete indexed lists $A$ and $B$ and determines, for every recovered fact $b\in B$, whether a semantically equivalent source fact exists in $A$. Paraphrases, reordered wording, and a shared core claim at different descriptive granularity count as equivalent; statements about different objects, events, or attributes, and statements that are merely inferentially related, do not. The model returns aligned arrays of zero-based $B$ and $A$ indices. Each $B$ fact may match at most one $A$ fact. We additionally scan returned pairs in order and discard later pairs that reuse an already matched $A$ fact, thereby enforcing one-to-one credit. Match outputs are cached using the first 16 hexadecimal characters of the source and recovered caption hashes.

\paragraph{Metrics and aggregation.}
For each video, $|M|$ is the number of retained pairs, $R=|M|/|A|$, $P=|M|/|B|$, and $\mathrm{CSRS}=2RP/(R+P)$. We also report Loss $=1-R$ and Drift $=1-P$. If both lists are empty, $R=P=\mathrm{CSRS}=1$; if only $A$ is empty, $R=1$, $P=0$, and CSRS $=0$; if only $B$ is empty, $R=0$, $P=1$, and CSRS $=0$. Fact density is the number of extracted facts per 100 words, counting each Chinese character as one word and whitespace-delimited English tokens as words. All reported configuration-level values are macro averages over the 33 per-video scores. The full pipeline yields 198 paired evaluations; fact extraction and semantic matching are cached separately to support incremental reruns without changing completed judgments.

\end{document}